\documentclass{article}
\usepackage{ijcai19}

\usepackage{times}
\usepackage{soul}
\usepackage{url}
\usepackage[hidelinks]{hyperref}
\usepackage[utf8]{inputenc}
\usepackage[small]{caption}
\usepackage{graphicx}
\usepackage{amsmath}
\usepackage{booktabs}
\usepackage{algorithm}
\usepackage{algorithmic}
\usepackage{graphicx}
\usepackage[super]{nth}
\title{Discriminating Spatial and Temporal Relevance in Deep Taylor Decompositions for Explainable Activity Recognition}

\author{
Liam Hiley$^{1,2}$\and
Harrison Taylor$^{1,2}$\and
Alun Preece$^{1,2}$\and
Yulia Hicks$^{1,3}$\and
David Marshall$^{1,2}$
\affiliations
$^1$Cardiff University, Crime and Security Research Institute\\
$^2$Cardiff University, School of Computer Science \& Informatics\\
$^3$Cardiff University, School of Engineering\\
\emails
\{HileyL,TaylorH23,PreeceAD,HicksYA,MarshallAD\}@cardiff.ac.uk, 
}

\begin{document}

\maketitle

\begin{abstract}
    Current techniques for explainable AI have been applied with some success to image processing. The recent rise of research in video processing has called for similar work in deconstructing and explaining spatio-temporal models. While many techniques are designed for 2D convolutional models, others are inherently applicable to any input domain. One such body of work, deep Taylor decomposition, propagates relevance from the model output distributively onto its input and thus is not restricted to image processing models. However, by exploiting a simple technique that removes motion information, we show that it is not the case that this technique is effective as-is for representing relevance in non-image tasks. We instead propose a discriminative method that produces a na\"ive representation of both the spatial and temporal relevance of a frame as two separate objects. This new \emph{discriminative relevance} model exposes relevance in the frame attributed to motion, that was previously ambiguous in the original explanation. We observe the effectiveness of this technique on a range of samples from the UCF-101 action recognition dataset, two of which are demonstrated in this paper.
\end{abstract}

\section{Introduction}
Recent success in solving image recognition problems can be attributed to the application to these problems of increasingly complex convolutional neural networks (CNNs) that make use of spatial convolutional feature extractors. This success has been closely followed by a call for explainability and transparency of these networks, which have inherently been regarded as black-boxes. Significant efforts have been made towards explaining decisions in image recognition tasks, with nearly all of these producing explanations through an image medium \cite{bach:lrp,simonyan:conv_visualising,baehrens:sensitivity,zhou:cam}.

Even more recently, the success in solving image recognition problems has been followed by the appearance of analogous video recognition models that make effective use of CNNs by extending the convolution dimensionality to be spatio-temporal or 3D \cite{ji:3dconv,carreira:quo,kensho-et-al:3dresnet}. Intuitively, the same methods that have been successful in explaining image models can be applied to video. Many of these methods, notably the popular deep Taylor decomposition \cite{montavon:deeptaylor}, function on any input without modification. However, the additional temporal dimension is not conceptually similar and, hence, exchangeable with the two spatial dimensions in the input. This is not accounted for by the model, which simply convolves the input in all dimensions in the same manner. This is reflected in explanations using image-based methods like deep Taylor. A pixel, or voxel in this case, is not marked whether it is temporally or spatially relevant, but on the basis of its combined spatio-temporal relevance.

By applying the deep Taylor method to a 3D convolutional network trained on the UCF-101 activity recognition dataset \cite{soomro:ucf101}, and additionally explaining the same class for each individual frame as a separate input, we effectively explain the spatial relevance of that frame. We show that by subtracting this from the original explanation, one can expose the underlying relevance attributed to motion in the frame. Thus we propose a new \emph{discriminative relevance model}, which reveals the relevance attributed to motion, that was previously hidden by the accompanying spatial component. 

\section{Related Work}
Inflating convolutional layers to 3D for video tasks was first explored in \cite{ji:3dconv}, in which the authors chose to optimise an architecture for the video task, rather than adapt one from an image problem.
Both \cite{carreira:quo} and \cite{kensho-et-al:3dresnet} have adapted large image classification models (Inception and ResNet respectively) to activity recognition tasks, such as \cite{kay:kinetics,soomro:ucf101}. Aside from the added dimensionality, these architectures are much the same as in image tasks, and intuitively find similar success in the spatio-temporal domain as they do in the spatial domain, achieving state-of-the-art performance. These models are as complex and black-box in nature as their 2D counterparts and as such the motivation to explain them also translates.

A variety of approaches have been attempted for explaining decisions made by deep neural networks. For example, in \cite{simonyan:conv_visualising} the authors propose feature visualisation for CNNs, in which the input images are optimised to maximally activate each filter in the CNN convolutional layers, following work in \cite{erhan:visualizing} on non-convolutional models. Local explanations, in the sense that they are local to a single input, explain the inputs contribution to the model decision using feature attribution; these have found much success in explaining deep image processing models. These methods in some way approximate the contribution to the models decision, most commonly in a supervised task, to its input variables, pixels or features at a higher level. This has been implemented in a number of ways, for example, through use of probability gradients \cite{baehrens:sensitivity}, global average pooling \cite{zhou:cam} and its generalisation to networks with hidden layers in \cite{selvaraju:gradcam}, or through local relevance based around a decision-neutral root point \cite{bach:lrp,montavon:deeptaylor,zhang:eb}. These works are all considered \emph{white-box} in that they use information from the model internal parameters, i.e., its weights and activations, in generating an explanation.

Layer-wise relevance propagation (LRP) rules, as defined in \cite{bach:lrp}, have found moderate success in explaining image recognition tasks. Multiple implementations and improvements have been made to these rules, with marginal winning probability (MWP) \cite{zhang:eb}, to our knowledge being the first implementation of the rules. Deep Taylor decomposition, an implementation of LRP by the original authors themselves has become very popular, and as a result of its input-domain agnosticism, has been applied to other domains outside of image recognition, including activity recognition \cite{srinivasan:videolrp}. It is for these reasons we choose the deep Taylor method as the exemplar technique for our proposed method

In addition to MWP, the authors in \cite{zhang:eb} also show that removing relevance for the dual of the signal improves the focus of the explanation. This contrastive MWP (cMWP) effectively removes relevance to all classes, by explaining all other outputs at the second logits layer, leaving only relevance contributing to the chosen output neuron. Our method is similar to cMWP, in that we make use of subtraction of separate LRP signals to remove unwanted relevance. However, we backpropagate both signals through the network fully before subtracting. Where the cMWP method removes relevance towards all classes from the explanation, our method removes relevance towards spatially salient features in the frame, such as edges and background objects.

Work on explainability methods outside of image tasks is still developing. Papers such as \cite{carreira:quo} use feature visualisation techniques to provide insight into the models they have trained, but to our knowledge \cite{srinivasan:videolrp} is still one of the only instances of an LRP based method applied to a video task. 
In this work, the difference between frames in relevance is highlighted by flattening the explanation block and plotting the overall relevance, which shows frames at certain points in an activity are more relevant overall.
Saliency tubes, as proposed in \cite{stergiou:tubes}, adapts the CAM technique of \cite{zhou:cam,selvaraju:gradcam} to localise salient motion in video frames. This method is the most similar to ours in that it highlights motion in 3D CNNs.

\section{Spatial and Temporal Relevance in 3D CNNs}
\subsection{3D CNNs}
3D CNNs extend the convolutional layers to the third dimension. In 3D CNNs, a sliding cube passes over a 3D block formed by the frames of the video stacked one on top of another, as opposed to a sliding 2D window passing over the image in 2D CNNs. This results not only in spatial features, but also features of motion, being learned. In the process of explaining the input, however, the relevance for the video is deconstructed into the original frames, which can be animated in the same manner as the input itself. Although the frames can be staggered, made transparent, and visualised as a block (see \cite{stergiou:tubes} for an example), the explanation is essentially viewed as a series of images; 
this is also the case with \cite{srinivasan:videolrp}.
In this manner it is impossible to distinguish the effect of the motion of the objects in the frame. At the same time, discerning whether a segment of the frame is considered relevant because of its shape and colour, its spatial information, or because of its  relationship to similar areas in the neighbouring frames, i.e., motion, can be important for explaining the decisions a CNN makes. 
While one can infer important frames from the approach in \cite{srinivasan:videolrp}, this does not address the issue of spatially relevant objects that aren't visible in other frames, nor does it necessarily localise the temporally important regions.

\subsection{Separating Spatial and Temporal Components through Discriminative Relevance}
Distinguishing spatial and temporal relevance of pixels (voxels) when using 3D CNNs is not always possible. A kernel does not necessarily focus on either spatial or temporal features, but in most cases will have components in both. As a result, decoupling spatial and temporal relevance is not intuitive given only the original explanation. Instead, the motion can be removed from the input video. By performing multiple additional forward passes, which each take as an input a single frame from the original video, the temporal information can effectively be excluded from the model's decision. The resulting video would depict a freeze frame at that instant, and thus the model would only have the spatial information in the scene to infer from. By doing this, we can build up an image of the purely spatial relevance in each frame. The required additional computation scales linearly with the size of the temporal dimension of an input sample. Intuitively, by downweighting relevance in the original explanation by the spatial relevance reconstructed from each frame's explanation, what is left will be the relevance based on motion.

\section{Implementation}

\begin{figure*}[tb]
    \centering
    \includegraphics[width=14cm]{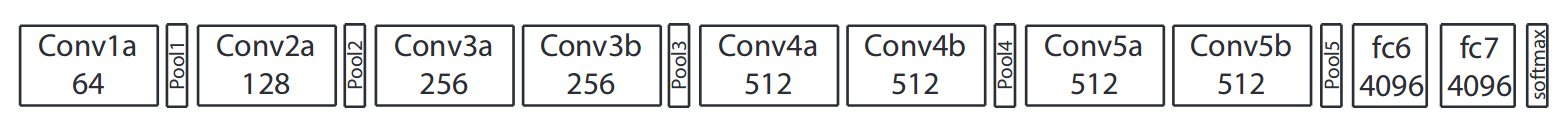}
    \caption{The chosen 3D architecture}
    \label{fig:c3d}
\end{figure*}

All work presented in this article is implemented in Python, using the deep learning platform \href{https://pytorch.org}{PyTorch}, with the autograd functionality modified to apply LRP rules during backpropagation in place of the normal gradient calculation.

\subsection{The \texttt{explain} PyTorch extension}
The PyTorch \texttt{autograd} library provides the basis on which symbolic graphs are created. The \texttt{Function} class can be extended to create mathematical operations with custom gradient functions. We adapted this library to work with LRP techniques as the unofficial \texttt{explain} library\footnote{This code is freely available at https://github.com/liamhiley/torchexplain}. On a forward pass, a model using the \texttt{explain} Function class will act in the same way as a PyTorch model using the \texttt{autograd} Function class. Weights are also loaded into the model in the same way as PyTorch models. The functionality differs on the backwards pass.
The \texttt{autograd} library allows for custom backwards functions to be defined. Through this feature, we implemented convolutional, batch norm, pooling and linear layers whose gradient functions instead propagate relevance. In this way, we can generate an LRP explanation for a model given an input, by performing a forward pass, and then backpropagating the relevance, beginning at the chosen output neuron, back onto the input sample.

\subsection{Model}
The model is a 3D CNN following the C3D design from \cite{tran:spatiotemporal}, the architecture for which is shown in Figure~\ref{fig:c3d}. The code is adapted from an implementation available at \href{https://github.com/jfzhang95/pytorch-video-recognition}{https://github.com/jfzhang95/pytorch-video-recognition}.
We fine-tuned the pretrained weights made available, to 75\% validation accuracy on the UCF-101 dataset.

\subsection{Deep Taylor Decomposition}
Our implementation of deep Taylor decomposition follows the most up-to-date version found on \href{https://github.com/albermax/innvestigate}{https://github.com/albermax/innvestigate}, a repository maintained by one of the lead authors on the LRP and deep Taylor papers.
The implementation is summarised as follows:
\begin{itemize}
    \item As in the original deep Taylor paper, ReLU nonlinearities simply pass on relevance, without modifying it in any way.
     $$
        R_{k} = R_{j}
     $$
    \item Relevance for pooling layers is generated by multiplying the input to the layer by the relevance w.r.t that input, calculated using regular backpropagation rules for pooling layers.
    For max-pooling:
     $$
        R_{k} = \delta_{j}^{k}R_{j}
     $$
    Where $\delta_{j}^{k}$ is a mask of whether the neuron $k$ was selected by the pooling kernel $j$.
    For average-pooling:
     $$
        R_{k} = \frac{R_{j}}{N_{j}}
     $$
    Where $N_{j}$ is the number of values in the pooling kernel $j$.
    \item The convolutional layers use the $\alpha\beta$ relevance rule, which focuses explanations by injecting some negative relevance.
     $$
        R_{i}=\sum_{j}(\alpha\frac{z_{ij}^{+}}{\sum_{i}z_{ij}^{+}+b_{j}^{+}}-\beta\frac{z_{ij}^{-}}{\sum_{i}z_{ij}^{-}+b_{j}^{-}})R_{j}
     $$
    \item The first convolutional layer (for which the input is the sample) uses the $z_{\beta}$ rule which makes use of the restricted input space (0 to 255 for pixel values) in finding a root point.
     $$
        R_{i}=\sum_j\frac{z_{ij}-l_{i}w_{ij}^{+}-h_{i}w_{ij}^{-}}{\sum_{i'}z_{i'j}-l_{i}w_{i'j}^{+}-h_{i}w_{i'j}^{-}}R_{j}
     $$
\end{itemize}

\subsection{Padding}
Without the use of adaptive pooling, the single frame inputs would be too small to pass through the network. Still, this would result in more of the spatial information for the frame conserved than originally as part of the video. Instead, we padded the frame to the same size as the input video. We chose to pad the input by repeating the frame $n$ times, where $n$ a typical sample size-number, rather than use zero padding, which would create false temporal information through the near-instant change from all pixels to black. This is supported by the findings in \cite{hooker:roar} where a similar issue arose when quantifying the accuracy of feature attribution techniques like LRP by replacing relevant pixels with zero-black pixels.

\section{Results}

\begin{figure*}[tb]
    \centering
    \includegraphics[width=5.5cm]{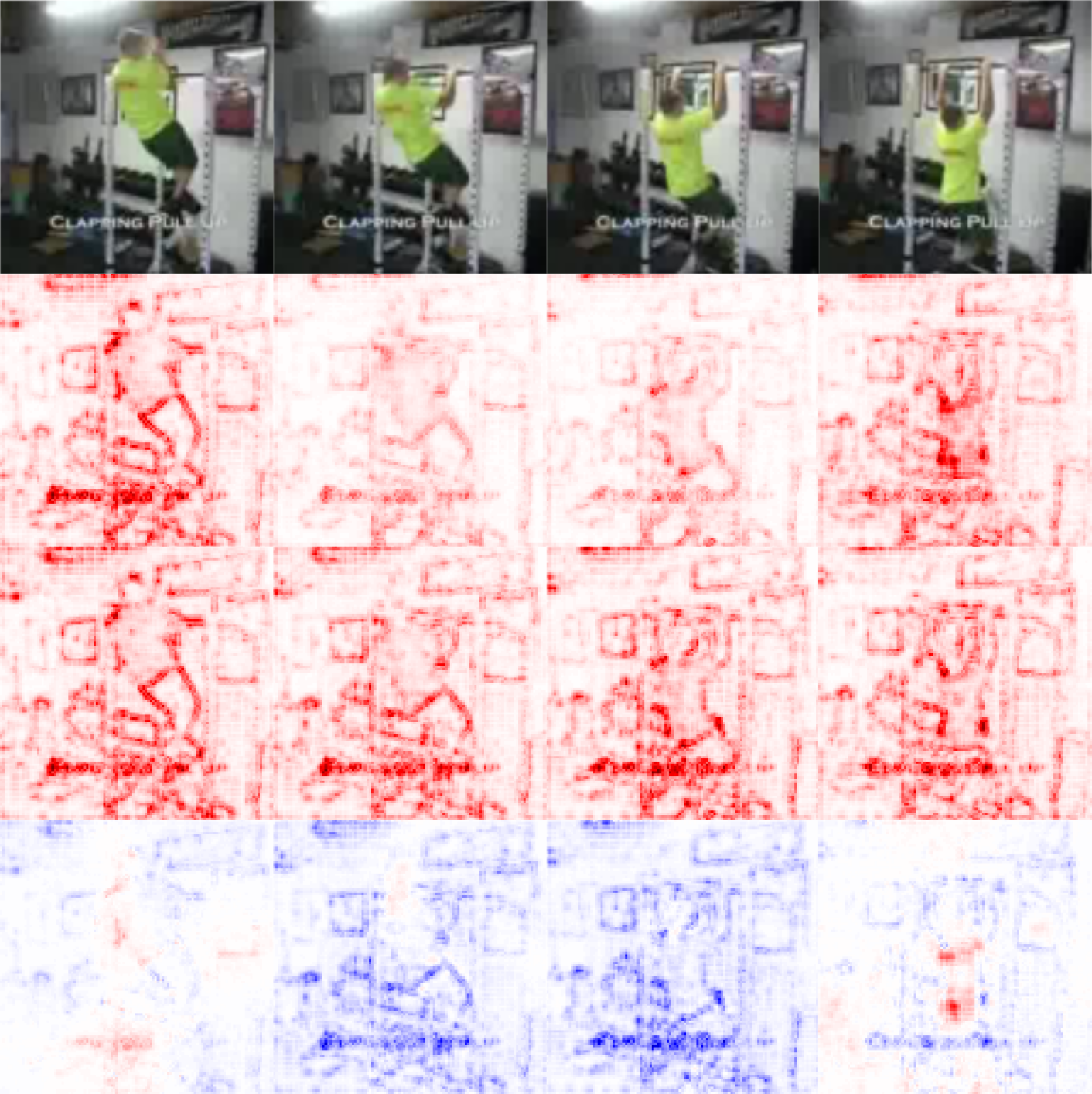}
    \hspace{.45cm}
    \includegraphics[width=5.5cm]{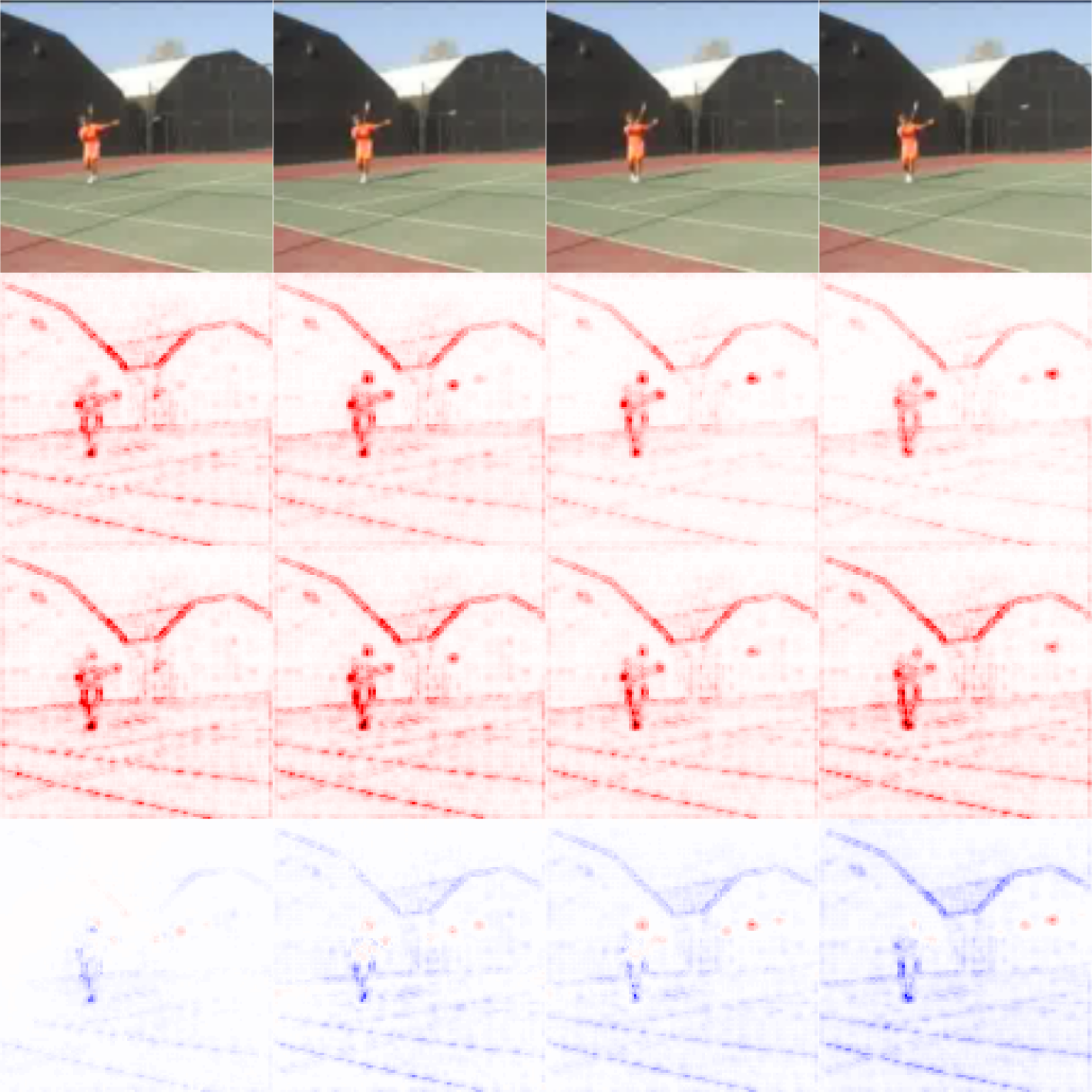}
    \caption{Left: The \nth{1}, \nth{6}, \nth{10} and \nth{16} frames respectively, from a 16-frame sample. Right: The \nth{1}, \nth{2}, \nth{3} and \nth{4} frames respectively, from a 16-frame sample.
    \nth{1} row: the original frames; \nth{2} row: the deep Taylor explanation for the sample; \nth{3} row: the spatial-only deep Taylor explanation for each frame; \nth{4} row: the remaining temporal explanation after subtracting (\nth{3}) from (\nth{2}). Red is positive relevance, blue is negative relevance, white is no relevance.}\label{fig:pullups}
\end{figure*}

In this section we show the result of subtracting spatial relevance from a deep Taylor explanation of an input video, featuring a person performing pull-ups on a bar. The three different explanations can be seen in Figure~\ref{fig:pullups}. The frames shown are processed, and explained as a three dimensional block, and displayed as two dimensional slices. 

In the original relevance (Figure \ref{fig:pullups}: Row 2), most of the scene is marked relevant, with heavy focus on edges. This observation is reinforced by the spatial relevance (Figure \ref{fig:pullups}: Row 3), which in the absence of temporal information, highlights edges much more heavily. 
The agreement between the spatial and original explanations demonstrates the ambiguity in 3D explanations with gradient-based techniques like deep Taylor. It is unclear even with the spatial explanation as a reference, what in the scene is relevant for its motion, as every object is to a degree marked relevant. The difference becomes clearer with the inclusion of the temporal explanation.
Subtracting the spatial explanation from the original explanation, shows a large amount of remaining relevance in the core of the man's body and his head. The relevance in the background, the metal frame and the video watermark are all negative as a result, suggesting they are all highly spatially relevant. This effect is displayed for the beginning, end, and two intermediate frames of the sample. The bulk of the temporal relevance is found at the highest and lowest points of the exercise, but is absent from the intermediate frames. This suggests that the motion at key moments of the activity are at the lowest and highest points of the pull up, possibly due to the sharp change in movement, as at these points both the lowering and raising of the body are observed. This information is much more difficult to infer from the original explanation, where the \nth{1} and \nth{16} frames are overall much more heavily red (or relevant).
In this example, key frames as well as salient regions are highlighted for motion. In the second example, of a person serving a tennis ball, the activity is observed at a more fine-grained framerate. Likely because of this the relevance of motion is more regular over the 4 neighbouring frames, when compared to the sparsely-sampled frames of the pull ups. This serves more to highlight temporally relevant objects in the scene. Specifically, the tennis ball and the person's upper body, where the swinging motion originates from. Again, in the original explanation this information is not clear. In fact, the ball and upper torso are relatively indistinguishable from spatially relevant features like the lawn and the building in the background. As seen in the spatial explanation, the relevance of these regions has a spatial component as well.
%

Also interesting is the change in prediction by the model when given only spatial information. For the \nth{1}, \nth{6}, \nth{10} and \nth{16} frame-only inputs, the model predicted Wall Push-Ups, Golf Swing, Clean And Jerk, and Clean And Jerk again, respectively. While only two samples are illustrated in this paper, similar results were observed for other test samples in the UCF-101 dataset indicating that these examples are not anomalous, and the method is general. 
The evidence that our method is an approximation is twofold. Firstly, the inequality between the sum of the spatial and temporal relevance, and the original relevance shows that the former are not true fractions of the latter. Furthermore, the fact that the spatial relevance for the non-dominant class (Pull Ups, when the model has decided Clean And Jerk) is more than the same frames explained as the dominant class (Pull Ups, when the model had decided Pull Ups) supports this.

\section{Conclusion}
In this paper we have introduced a new use case, for separating and visualising the spatial and temporal components of an explanation by deep Taylor decomposition, for a spatio-temporal CNN, that is easy to implement and takes relatively little extra computational cost. By exploiting a simple method of removing motion information from video input, we have essentially generated a negative mask that can be applied to an explanation that will remove the spatial component in the relevance. The resulting explanation provides much more insight into the salient motion in the input than the general relevance, which we show is noisy with misleading spatial relevance, i.e., most edges in the frame.

While we expose an unsuitability in the current implementation of the deep Taylor method, for inputs with non-exchangeable dimensions, this work is ongoing. In the future, it will be necessary to formalise a method for exposing the true spatial relevance in the frame, as opposed to an approximation such as our method.

\section*{Acknowledgements} 
This research was sponsored by the U.S. Army Research Laboratory and the UK Ministry of Defence under Agreement Number W911NF-16-3-0001. The views and conclusions contained in this document are those of the authors and should not be interpreted as representing the official policies, either expressed or implied, of the U.S. Army Research Laboratory, the U.S. Government, the UK Ministry of Defence or the UK Government. The U.S. and UK Governments are authorised to reproduce and distribute reprints for Government purposes notwithstanding any copyright notation hereon


\end{document}